\documentclass[11pt, letterpaper]{article}
\usepackage{acl2012}
\usepackage{times}
\usepackage{url}
\usepackage{latexsym}
\usepackage{xcolor}
\usepackage{graphicx}
\usepackage{verbatim} 
\usepackage{multirow} 
\usepackage{amssymb,amsmath}
\usepackage{balance}
\usepackage{color}

\usepackage[utf8]{inputenc}
\usepackage{graphicx}
\usepackage{amssymb,amsmath}
\usepackage[T1]{fontenc}

\newenvironment{itemizesquish}{\begin{list}{\labelitemi}{\setlength{\itemsep}{0em}\setlength{\labelwidth}{0.5em}\setlength{\leftmargin}{\labelwidth}\addtolength{\leftmargin}{\labelsep}}}{\end{list}}

\title{New Alignment Methods for Discriminative Book Summarization\\ \textcolor{blue}{Work in Progress}}
\author{David Bamman \and Noah A.~Smith\\
 School of Computer Science\\
  Carnegie Mellon University \\
Pittsburgh, PA 15213, USA \\
  {\tt \{dbamman,nasmith\}@cs.cmu.edu} 
  \\}

\begin{document}
\maketitle

\begin{abstract}
We consider the unsupervised alignment of the full text of a book with a human-written summary.  This presents challenges not seen in other text alignment problems, including a disparity in length and, consequent to this, a violation of the expectation that individual words and phrases \emph{should} align, since large passages and chapters can be distilled into a single summary phrase.  We present two new methods, based on hidden Markov models, specifically targeted to this problem, and demonstrate gains on an extractive book summarization task.  While there is still much room for improvement, unsupervised alignment holds intrinsic value in offering insight into what features of a book are deemed worthy of summarization.
 \end{abstract}

\section{Introduction}

The task of \emph{extractive summarization} is to select a subset of sentences from a source document to present as a summary.   Supervised approaches to this problem make use of training data in the form of source documents paired with existing summaries\ \cite{Marcu:1999:ACL:312624.312668,Osborne:2002:UME:1118162.1118163,Jing:1999:DHS:312624.312666,ceylan:the}. These methods learn what features of a source sentence are likely to result in that sentence appearing in the summary; for news articles, for example, strong predictive features include the position of a sentence in a document (earlier is better), the sentence length (shorter is better), and the number of words in a sentence that are among the most frequent in the document.

Supervised discriminative summarization relies on an alignment between a source document and its summary.  For short texts and training pairs where a one-to-one alignment between source and abstract sentences can be expected, standard techniques from machine translation can be applied, including word-level alignment\ \cite{Brown:1990:SAM:92858.92860,Vogel:1996:HWA:993268.993313,Och:2003:SCV:778822.778824} and longer phrasal alignment\ \cite{Daume:2005:IWP:1110825.1110829}, especially as adapted to the monolingual setting\ \cite{quirk-brockett-dolan:2004:EMNLP}.  For longer texts where inference over all possible word alignments becomes intractable, effective approximations can be made, such as restricting the space of the available target alignments to only those that match the identity of the source word\ \cite{Jing:1999:DHS:312624.312666}.

\begin{figure}[htp*]
\begin{centering}
\includegraphics[scale=.6]{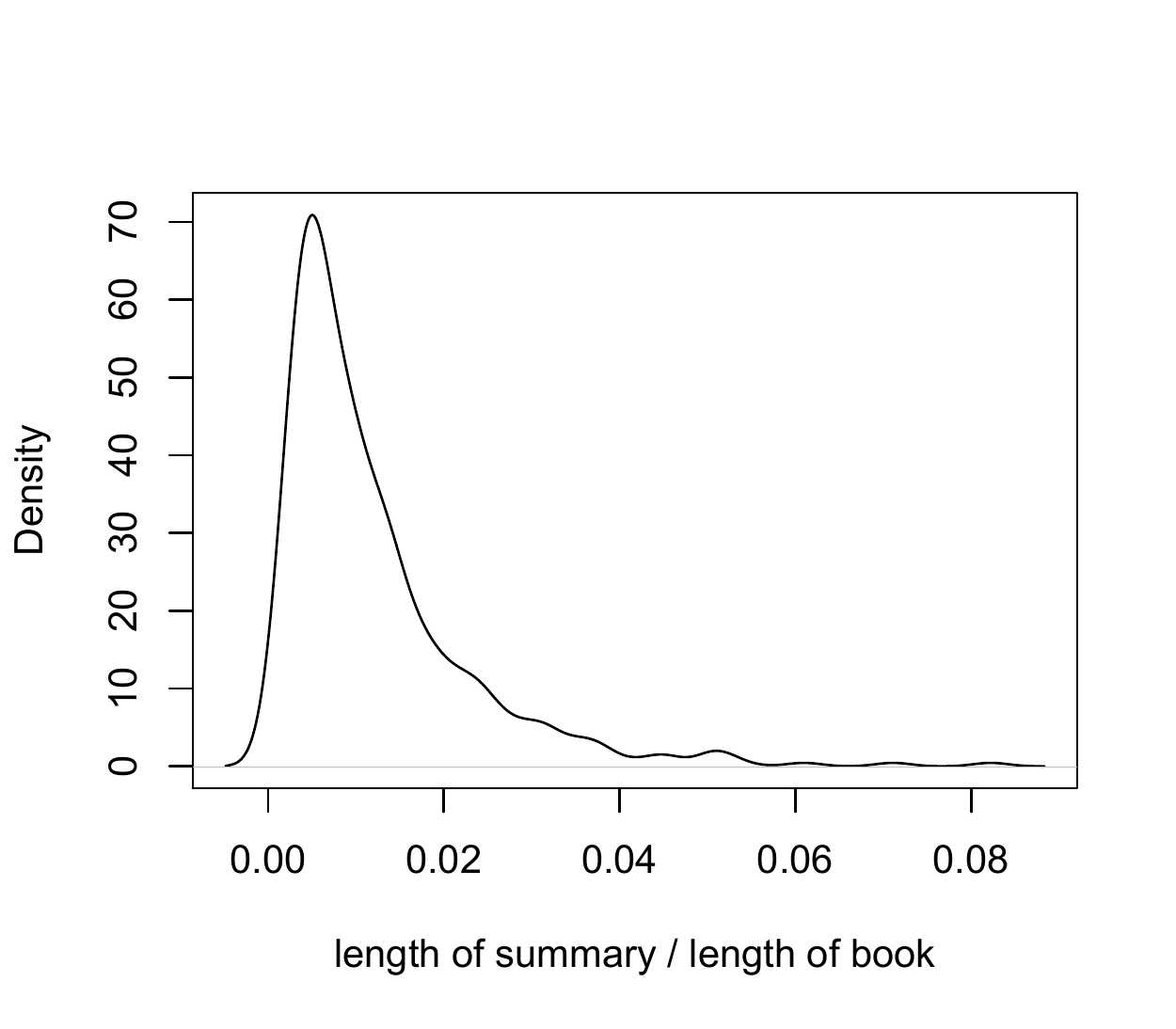}
\caption[]{Size disparity between summaries and full texts. Summaries average 1\% the size of the corresponding book. The mean is 0.012, with a $[5,95]$ quantile of $[0.002, 0.032$].}
\label{len}
\end{centering}
\end{figure}

The use of alignment techniques for book summarization, however, challenges some of these assumptions.  The first is the disparity between the length of the source document and that of a summary.  While the ratio between abstracts and source documents in the benchmark Ziff-Davis corpus of newswire\ \cite{Marcu:1999:ACL:312624.312668}  is approximately 12\% (133 words vs. 1,066 words), the length of a full-text book greatly overshadows the length of a simple summary.  Figure \ref{len} illustrates this with a dataset comprised of books from Project Gutenberg paired with plot summaries extracted from Wikipedia for a set of 439 books (described more fully in \S \ref{data} below).  The average ratio between a summary and its corresponding book is $1.2\%$.

This disparity in size leads to a potential violation of a second assumption: that we expect words and phrases in the source document to align with words and phrases in the target.  When the disparity is so great, we might rather expect that an entire paragraph, page, or even chapter in a book aligns to a single summary sentence. 

To help adapt existing methods of supervised document summarization to books, we present two alignment techniques that are specifically adapted to the problem of book alignment, one that aligns passages of varying size in the source document to sentences in the summary, guided by the unigram language model probability of the sentence under that passage; and one that generalizes the HMM alignment model of \newcite{Och:2003:SCV:778822.778824} to the case of long but sparsely aligned documents.

\section{Related Work}

This work builds on a long history of unsupervised word and phrase alignment originating in the machine translation literature, both for the task of learning alignments across parallel text\ \cite{Brown:1990:SAM:92858.92860,Vogel:1996:HWA:993268.993313,Och:2003:SCV:778822.778824,DeNero:2008:SAS:1613715.1613758} and between monolingual\ \cite{quirk-brockett-dolan:2004:EMNLP} and comparable corpora\ \cite{Barzilay:2003:SAM:1119355.1119359}.  For the related task of document/abstract alignment, we draw on work in document summarization\ \cite{Marcu:1999:ACL:312624.312668,Osborne:2002:UME:1118162.1118163,Daume:2005:IWP:1110825.1110829}.  Past approaches to fictional summarization, including both short stories\ \cite{DBLP:journals/coling/KazantsevaS10} and books\ \cite{mihalcea-ceylan:2007:EMNLP-CoNLL2007}, have tended toward non-discriminative methods; one notable exception is Ceylan\ \shortcite{ceylan}, which applies the Viterbi alignment method of Jing and McKeown\ \shortcite{Jing:1999:DHS:312624.312666} to a set of 31 literary novels.

\section{Methods}

We present two methods, both of which involve estimating the parameters of a hidden Markov model (HMM).  The HMMs differ in their definitions of states, observations, and parameterizations of the emission distributions.  We present a generic HMM first, then instantiate it with each of our two models, discussing their respective inference and learning algorithms in turn.

Let $\mathcal{S}$ be the set of hidden states and $K = |\mathcal{S}|$.  An observation sequence $\boldsymbol{t} = \langle t_1, \ldots, t_n\rangle$, each $t_\ell \in \mathcal{V}$, is assigned probability:
\begin{align}
p(\boldsymbol{t} \mid n) & = \sum_{\boldsymbol{z} \in \mathcal{S}^{n}} \pi_{z_1} \left(\prod_{\ell=1}^{n} \eta_{z_\ell, t_\ell} \gamma_{z_{\ell}, z_{\ell+1}} \right) 
\end{align}
\noindent where $\boldsymbol{z}$ is the sequence of hidden states, $\boldsymbol{\pi} \in \Delta_K$ is the distribution over start states, and for all $s \in \mathcal{S}$, $\boldsymbol{\eta}_s \in \Delta_{|\mathcal{V}|}$ and $\boldsymbol{\gamma}_s \in \Delta_{K}$ are $s$'s emission and transition distributions, respectively.  Note that we avoid stopping probabilities by always conditioning on the sequence length.

\begin{table*}
\begin{center}
\begin{tabular}{lll}
& Passage model & Token model \\ \hline
states $\mathcal{S}$ & source document passages & source document tokens \\
 observations & summary sentences & summary tokens \\
transitions & by passage order difference  & by distance bin \\
emissions &  unigram distribution & lexical identity, synonyms \\
\hline
\end{tabular}
\end{center}
\caption{Summary of the passage model (\S\ref{se:passage-model}) and the token model (\S\ref{se:second-model}).}
\end{table*}

\subsection{Passage Model} \label{se:passage-model}

In the passage model, each HMM state corresponds to a contiguous passage in the source document.    The intuition behind this approach is the following: while word and phrasal alignment attempts to capture fine-grained correspondences between a source and target document, longer documents that are distilled into comparatively short summaries may instead have long, topically coherent passages that are summarized into a single sentence.  For example, the following summary sentence in a Wikipedia plot synopsis summarizes several long episodic passages in \emph{The Adventures of Tom Sawyer}:

\begin{quotation}
After playing hooky from school on Friday and dirtying his clothes in a fight, Tom is made to whitewash the fence as punishment all of the next day.
\end{quotation}

Our aim is to find the sequence of passages in the source document that aligns to the sequence of summary sentences.  Therefore, we identify each HMM state in $s \in \mathcal{S}$ with source document positions $i_s$ and $j_s$.  When a summary sentence $t_{\ell} = \langle t_{\ell,1}, \ldots, t_{\ell,T_{\ell}}\rangle$ is sampled from state $s$, its emission probability is defined as follows:
\begin{align}
\eta_{s, t_{\ell}} & = \prod_{k=1}^{T_\ell} \hat{p}_{\mathit{unigram}}(t_{\ell,k} \mid \boldsymbol{b}_{i_s:j_s}) \label{eq:emission}
\end{align}
where $\boldsymbol{b}_{i_s:j_s}$ is the passage in the source document from position $i_s$ to position $j_s$; again, we avoid a stop symbol by implicitly assuming lengths are fixed exogenously.  The unigram distribution $\hat{p}_{\mathit{unigram}}(\cdot \mid \boldsymbol{b}_{i_s:j_s})$ is estimated directly from the source document passage $\boldsymbol{b}_{i_s:j_s}$.  

The transition distribution from state $s \in \mathcal{S}$, $\boldsymbol{\gamma}_s$ is operationalized following the HMM word alignment
formulation of \newcite{Vogel:1996:HWA:993268.993313}.   The transition events between ordered pairs of states are binned by the difference in two passages' ranks within the source document.\footnote{These ranks are fixed; our inference procedure does not allow passages to overlap or to ``leapfrog'' over each other across iterations.}  We give the formula for
relative frequency estimation of the transition distributions:
\begin{align}
\gamma_{s, s'}   & = \frac{c(s' - s)}{\sum_{s'' \in \mathcal{S}} c(s - s'')}
\end{align}
where $c(\cdot)$ denotes the count of jumps of a particular length, measured as the distance between the rank order of two passages within a document; the count of a jump between passage $10$ and passage $13$ is the same as that between passage $21$ and $24$; namely, $c(3)$. Note that this distance is signed, so that the distance of a backwards jump from passage 13 to passage 10 ($-3$) is not the same as a jump from 10 to 13 ($3$).

\begin{figure*}[htp]
\begin{centering}
\includegraphics[scale=1]{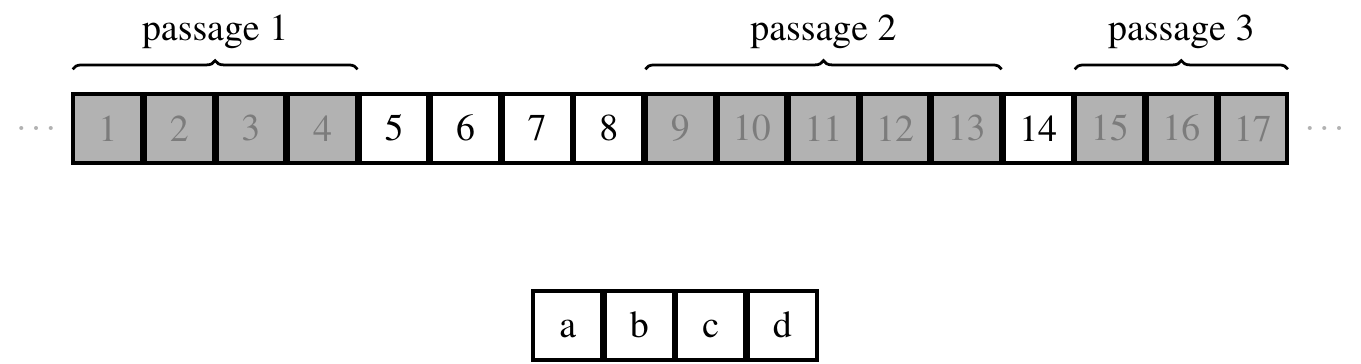}
\caption[]{Illustration of the passage HMM.   HMM states correspond to passages in the source document (top); each emission is a summary sentence (bottom).} 
\label{chain}
\end{centering}
\end{figure*}

The HMM states' spans are constrained not to overlap with each other, and they need not cover the source document.  Because we do not know the boundary positions for states in advance, we must estimate them alongside the traditional HMM parameters.
Figure \ref{chain} illustrates this scenario with a sequence of 17 words in the source document ($[1\ldots 17]$) and 4 sentences in the target summary ($\{a,b,c,d\}$).  In this case, the states correspond to $[1\ldots 4], [9\ldots 13]$, and $[15\ldots17]$.

\subsubsection{Inference} \label{se:passage-inference}

Given a source document $\boldsymbol{b}$ and a target summary $\boldsymbol{t}$, our aim is to infer the most likely passage $z_\ell$ for each sentence $t_\ell$.  This depends on the parameters ($\boldsymbol{\pi}$, $\boldsymbol{\eta}$, and $\boldsymbol{\gamma}$) and the passages associated with each state, so we estimate those as well, seeking to maximize likelihood.  Our approach is an EM-like algorithm \cite{dempster:1977}; after initialization, it iterates among three steps:
\begin{itemize}
\item \emph{E-step.} Calculate $p(\boldsymbol{t})$ and the posterior distributions $q(z_k \mid \boldsymbol{t})$ for each sentence $t_k$.  This is done using the forward-backward algorithm.
\item \emph{M-step.} Estimate $\boldsymbol{\pi}$ and $\boldsymbol{\gamma}$ from the posteriors, using the usual HMM M-step. 
\item \emph{S-step.} Sample new passages for each state.  The sampling distribution considers, for each state $s$, moving $i_s$ subject to the no-overlapping constraint and $j_s$, and then moving $j_s$ subject to the no-overlapping constraint and $i_s$ \cite{DeNero:2008:SAS:1613715.1613758}.  (See \S\ref{se:sampling} for more details.)
The emission distribution $\boldsymbol{\eta}_{s}$ is updated whenever $i_s$ and $j_s$ change, through Equation~\ref{eq:emission}. 
\end{itemize}

For the experiments described in section \ref{experiments}, each source document is initially divided into $K$ equal-length passages ($K=100$), from which initial emission probabilities are defined; $\boldsymbol{\pi}$ and $\boldsymbol{\gamma}$ are both initialized to uniform distribution.  Boundary samples are collected once for each iteration, after one E step and one M step, for a total of 500 iterations.

\subsubsection{Sampling chunk boundaries} \label{se:sampling}

During the S-step, we sample the boundaries of each HMM state's passage, favoring (stochastically) those boundaries that make the observations more likely.  We expect that, early on, most chunks will be radically reduced to smaller spans that match closely the target sentences aligned to them with high probability.  Over subsequent iterations, longer spans should be favored when adding words at a boundary offsets the cost of adding the non-essential words between the old and new boundary.

A greedy step---analogous to the M-step use to estimate parameters---is one way to do this:  we could, on each S-step, move each span's boundaries to the positions that maximize likelihood under the revised language model.  Good local choices, however, may lead to suboptimal global results, so we turn instead to sampling.  Note that, if our model defined a marginal distribution over passage boundary positions in the source document, this sampling step could be interpreted as part of a Markov Chain Monte Carlo EM algorithm \cite{Wei1990Monte}.  As it is, we do not have such a distribution; this equates to a fixed uniform distribution over all valid (non-overlapping) passage boundaries.

The implication is that the probability of a particular state $s$'s passage's start- or end-position is proportional to the probability of the observations generated given that span.  Following any E-step, the assignment of observations to $s$ will be fractional.  This means that the likelihood, as a function of particular values of $i_s$ and $j_s$, depends on all of the sentences in the summary:
\begin{align}
L(i_s, j_s) &= \prod_{\ell = 1}^n  \eta_{s,t_\ell}^{q(z_\ell = s \mid \boldsymbol{t}) } \label{like} \\
& =
\prod_{\ell = 1}^n   \left(\prod_{k=1}^{T_\ell } \hat{p}_{\mathit{unigram}}(t_{\ell,k} \mid \boldsymbol{b}_{i_s:j_s})\right) ^{q(z_\ell = s \mid \boldsymbol{t})} 
\nonumber
\end{align}

For example, in Figure~\ref{chain}, the start position of the second span (word 9) might move anywhere from word 5 (just past the end of the previous span) to word 12 (just before the end of its own span, $j_s = 12$).  Each of the values should be sampled with probability proportional to Equation~\ref{like}, so that the sampling distribution is:
\begin{align*}
\frac{1}{ \sum_{i = 5}^{12} L(i,12)} \langle L(5, 12), L(6, 12), \ldots, L(12,12) \rangle 
\end{align*}

Calculating $L$ for different boundaries requires recalculating the emission probabilities $\eta_{s,t_{\ell}}$ as the language model changes.
We can do this efficiently (in linear time) by decomposing the language model probability.  Here we represent a state $s$ by its boundary positions in the source document, $i:j$, and we use the relative frequency estimate for $\hat{p}_{\mathit{unigram}}$.
\begin{align}
\log \eta_{i:j, t_\ell} &= \sum_{k=1}^{T_\ell} \log \frac{\mathit{freq}(t_{\ell,k} ; \boldsymbol{b}_{i:j})}{j-i+1} \\
&= -T_{\ell} \log(j-i+1) + \sum_{k=1}^{T_\ell} \log \mathit{freq}(t_{\ell,k} ; \boldsymbol{b}_{i:j})
\end{align}
Now consider the change if we remove the first word from $s$'s passage, so that its boundaries are $[i+1, j]$.  Let $b_i$ denote the source
document's word at position $i$. $\log \eta_{i+1:j, t_\ell} =$
\begin{align}
\lefteqn{ -T_{\ell} \log(j-i) + \sum_{k=1}^{T_\ell} \log \mathit{freq}(t_{\ell,k} ; \boldsymbol{b}_{i+1:j}) } \nonumber \\
&=  \log \eta_{i:j, t_\ell} 
+ \mathit{freq}(b_i ; t_\ell) \log \frac{\mathit{freq}(b_i ; \boldsymbol{b}_{i:j}) - 1 }{\mathit{freq}(b_i ; \boldsymbol{b}_{i:j})}  \nonumber \\
& \  \ + T_\ell \log \frac{j-i+1}{j-i}
\end{align}
\noindent This recurrence is easy to solve for all possible left boundaries (respecting the no-overlap constraints) if we keep track of the word frequencies in each span of the source document---something we must do anyway to calculate $\hat{p}_{\mathit{unigram}}$.  A similar recurrence holds for the right boundary of a passage.

\begin{figure}[htp]
\begin{centering}
\includegraphics[scale=.6]{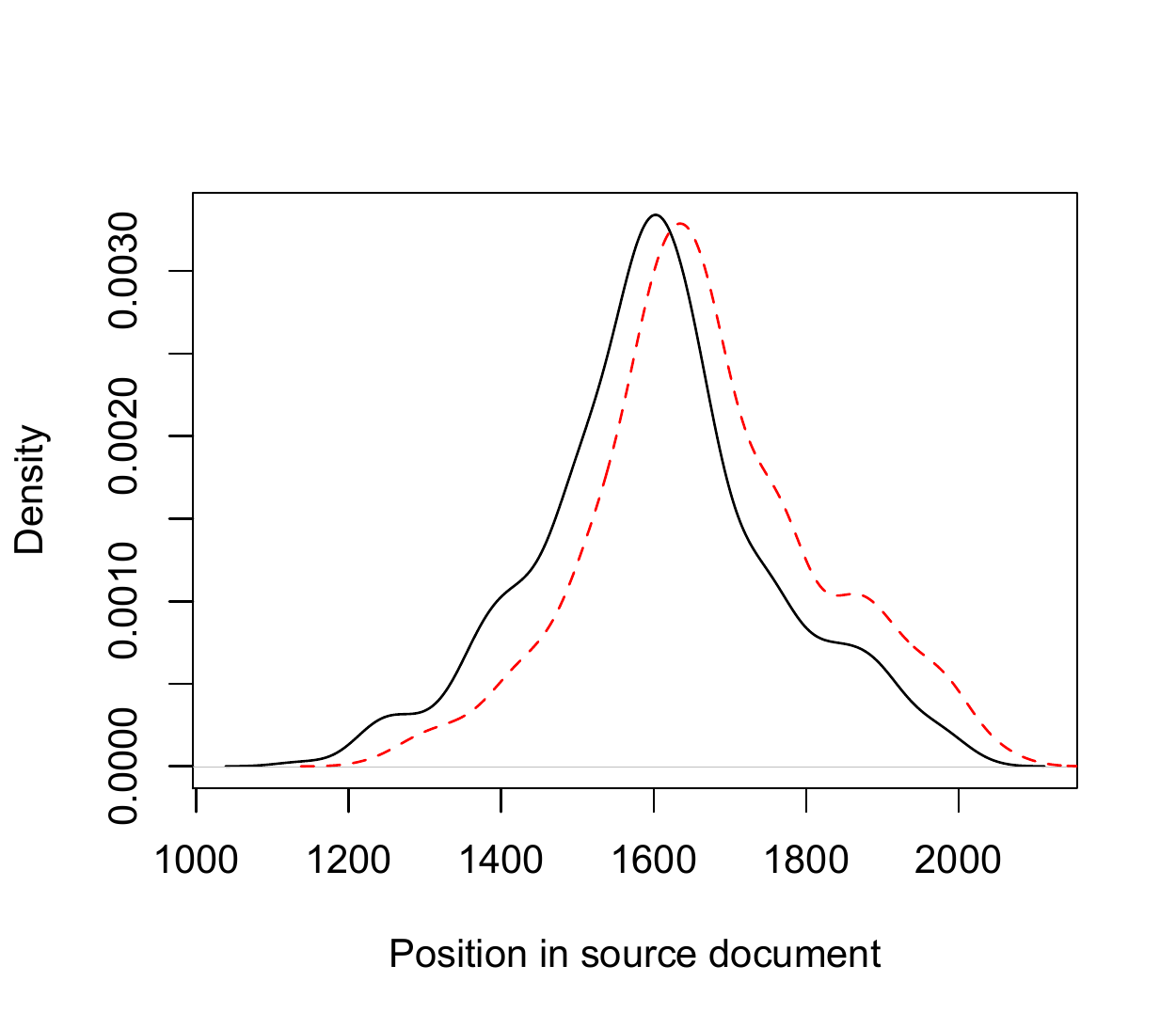}
\caption[]{Density plot of accumulated samples for one passage HMM state, in \emph{Heart of Darkness}.  The left boundary is shown in black and solid, the right boundary in red and dashed.}
\label{hd}
\end{centering}
\end{figure}

Figure \ref{hd} illustrates the result of this sampling procedure on the start and end positions for a single source passage in \emph{Heart of Darkness}.  After 500 iterations, the samples can be seen to fluctuate over a span of approximately 600 words; however, the modes are relatively peaked, with the most likely start position at 1613, and the most likely end position at 1660 (yielding a span of 47 words).

\subsection{Token Model} \label{se:second-model} 

\newcite{Jing:1999:DHS:312624.312666} introduced
an HMM whose states correspond to tokens in the source document.  The observation is the sequence of target summary tokens (restricting to those types found in the source document).  The emission probabilities are fixed to be one if the source and target words match, zero if they do not.  Hence each instance of $v \in \mathcal{V}$ in the target summary is assumed to be aligned to an instance of $v$ in the source.  The transition parameters were fixed manually to simulate a ranked set of transition types (e.g., transitions within the same sentence are more likely than transitions between sentences).  No parameter estimation is used; the Viterbi algorithm is used to find the most probable alignment. The allowable transition space is bounded by $F^2$, where $F$ is the frequency of the most common token in the source document.  The resulting model is scalable to large source documents \cite{ceylan:the,ceylan}.

One potential issue with this model  is that it lacks the concept of a null source, not articulated in the original HMM alignment model of \newcite{Vogel:1996:HWA:993268.993313} but added by \newcite{Och:2003:SCV:778822.778824}.  Without such a null source, every word in the summary must be generated by some word in the source document.  The consequence of this decision is that a Viterbi alignment over the summary must pick a perhaps distant, low-probability word in the source document if no closer word is available.  Additionally, while the choice to enforce lexical identity constrains the state space, it also limits the range of lexical variation captured.

Our second model extends Jing's approach  in three ways.

First, we introduce parameter inference to learn the values of start probabilities and transitions that maximize the likelihood of the data, using the EM algorithm.
We operationalize the transition probabilities again following \newcite{Vogel:1996:HWA:993268.993313}, but constrain the state space by only measuring transititions between fixed bucket lengths, rather than between the absolute position of each source word.  The relative frequency estimator for transitions is:
\begin{align}
\gamma_{s, s'} &=  {c(b(s' - s )) \over \sum_{s'' \in \mathcal{S}} c(b(s'' - s))}
\end{align}
As above, $c(\cdot)$ denotes the count of an event, and here $b(\cdot)$ is a function that transforms the difference between two token positions into a coarser set of bins (for example, $b$ may transform a distance of 0 into its own bin, a distance of $+1$ into a different bin, a distance in the range $[+2,+10]$ into a third bin, a difference of $[-10,-2]$ into a fourth, etc.).  Future work may include dynamically learning optimizal bin sizes, much as boundaries are learned in the passage HMM.

Second, we introduce the concept of a null source that can generate words in the target sentence. In the sentence-to-sentence translation setting, for a source sentence that is $m$ words long, \newcite{Och:2003:SCV:778822.778824} add $m$ corresponding NULL tokens, one for each source word position, to be able to adequately model transitions to, from and between NULL tokens in an alignment.  For a source \emph{document} that is ca.~100,000 words long, this is clearly infeasible (since the complexity of even a single round of forward-backward inference is $O(m^2n)$, where $n$ is the number of words in the target summary $\boldsymbol{t}$). However, we can solve this problem by noting that the transition probability as defined above is not measured between individual words, but rather between the positions of coarser-grained chunks that contain each word; by coarsing the transitions to model the jump between a fixed set of $B$ bins (where $B \ll m$), we effectively only need to add $B$ null tokens, making inference tractable.  As a final restriction, we disallow transitions between source state positions $i$ and $j$ where $|i -j| > \tau$.  In the experiments described in section \ref{experiments}, $\tau=1000$.

Third, we expand the emission probabilities to allow the translation of a source word into a fixed set of synonyms (e.g., as derived from Roget's Thesaurus.\footnote{\url{http://www.gutenberg.org/ebooks/10681}})   This expands the coverage of important lexical variants while still constraining the allowable emission space to a reasonable size.  All synonyms of a word are available as potential ``translations''; the exact translation probability (e.g., $\eta_{\textrm{purchase}, \textrm{buy}}$) is learned during inference.

\section{Experiments}
\label{experiments}

To evaluate these two alignment methods and compare with past work, we evaluate on the downstream task of extractive book summarization.

\subsection{Data}
\label{data}

The available data includes 14,120 book plot summaries extracted from the November 2, 2012 dump of English-language Wikipedia\footnote{\url{http://dumps.wikimedia.org/enwiki/}} and 31,393 English-language books from Project Gutenberg.\footnote{\url{http://www.gutenberg.org}}.  We restrict the book/summary pairs to only those where the full text of the book contains at least 10,000 words and the paired abstract contains at least 100 words (stopwords and punctuation excluded).  This results in a dataset of 439 book/summary pairs, where the average book length is 43,223 words, and the average summary length is 369 words (again, not counting stopwords and punctuation).

The ratio between summaries and full books in this dataset is approximately 1.2\%, much smaller than that used in previous work for any domain, even for past work involving literary novels: Ceylan\ \shortcite{ceylan:the} makes use of a collection of 31 books paired with relatively long summaries from \mbox{SparkNotes}, CliffsNotes and GradeSaver, where the average summary length is 6,800 words.  We focus instead on the more concise case, targeting summaries that distill an entire book into approximately 500 words.

\subsection{Discriminative summarization}

We follow a standard approach to discriminative summarization.  All experiments described below use 10-fold cross validation, in which we partition the data into ten disjoint sets, train on nine of them and then test on the remaining held-out partition.  Ten evaluations are conducted in total, with the reported accuracy being the average across all ten sets. 
First, all source books and paired summaries in the training set are aligned using one of the three unsupervised methods described above (Passage HMM, Token HMM, Jing 1999).

Next, all of the sentences in the source side of the book/summary pairs are featurized; all sentences that have been aligned to a sentence in the summary are assiged a label of 1 (appearing in summary) and 0 otherwise (not appearing in summary).  Using this featurized representation, we then train a binary logistic regression classifier with $\ell_2$ regularization on the training data to learn which features are the most indicative of a source sentence appearing in a summary.  Following previous work, we devise sentence-level features that can be readily computed in comparison both with the document in which the sentence in found, and in comparison with the collection of documents as  whole\ \cite{Yeh:2005:TSU:1041178.1041184,Shen:2007:DSU:1625275.1625736}.  All feature values are binary:

\begin{itemize}
\item Sentence position within document, discretized into membership in each of ten deciles. (10 features.)
\item Sentence contains a salient name.  We operationalize ``salient name'' as the 100 capitalized words in a document with the highest TF-IDF score in comparison with the rest of the data; only non-sentence-initial tokens are used for calculate counts.  (100 features.)
\item Contains lexical item $x$ ($x \in$ most frequent 10,000 words).  This captures the tendency for some actions, such as \emph{kills, dies} to be more likely to appear in a summary.  (10,000 features.)
\item Contains the \emph{first} mention of lexical item $x$ ($x \in$ most frequent 10,000 words).   (10,000 features.)
\item Contains a word that is among the top [1,10], [1,100], [1,1000] words having the highest TF/IDF scores for that book. (3 features.)
\end{itemize}

With a trained model and learned weights for all features, we next featurize each sentence in a test book according to the same set of features described above and predict whether or not it will appear in the summary.  Sentences are then ranked by probability and the top sentences are chosen to create a summary of 1,000 words.  To create a summary, sentences are then ordered according to their position in the source document.

\section{Evaluation}

Document summarization has a standard (if imperfect) evaluation in the ROUGE score\ \cite{Lin:2003:AES:1073445.1073465}, which, as an $n$-gram recall measure, stresses the ability of the candidate summary to recover the words in the reference.   To evaluate the automatically generated summary, we calculate the ROUGE score between the generated summary and the held-out reference summary from Wikipedia for each book.
We consider both ROUGE-1, which measures the overlap of unigrams, and ROUGE-2, which measures bigram overlap.  For the case of a single reference translation, ROUGE-N is calculated as the following (where $w$ ranges over all unigrams or bigrams in the reference summary, depending on $N$, and $c(\cdot)$ is the count of the $n$-gram in the text).

\begin{align}
{ \sum_{w \in ref} \min(c(w_{ref}), c(w_{hyp})) \over \sum_{w \in ref} c(w_{ref}) } 
\end{align}

Figure \ref{rouge} lists the results of a 10-fold test on the 439 available book/summary pairs.  Both alignment models described above show a moderate improvement over the method of Jing et al.  For comparison, we also present a baseline of simply choosing the first 1,000 words in the book as the summary.

\begin{table}[ht]
\begin{centering}
\begin{tabular}{|r|c|c|}
\hline
Model& ROUGE-1&ROUGE-2 \\ \hline \hline
Block HMM&41.4&6.2\\ \hline
Word HMM&41.3&6.2\\ \hline
Jing 1999&40.7&6.0\\ \hline
First 1000&38.0&6.0\\ \hline

\end{tabular}
\caption{\label{rouge} ROUGE summarization scores.}
\end{centering}
\end{table}

How well does this method actually work in practice, however, at the task of generating summaries?  Manually inspecting the generated summaries reveals that automatic summarization of books still has great room for improvement, for all alignment methods involved.  Appendix A shows the sentences extracted as a summary for \emph{Heart of Darkness}.  

Independent of the quality of the generated summaries on held-out test data, one practical benefit of training binary log-linear models is that the resulting feature weights are \emph{interpretable}, providing a data-driven glimpse into the qualities of a sentence that make it conducive to appearing in human-created summary.  Table \ref{words} lists the 25 strongest features predicting inclusion in the summary (rank-averaged over all ten training splits).  The presence of a name in a sentence is highly predictive, as is its position at the beginning of a book (decile 0) or at the very end (decile 8 and 9).  The strongest lexical features illustrate the importance of a character's persona, particularly in their relation with others (\emph{father}, \emph{son}, etc.), as well as the natural importance of major life events (\emph{death}).  The importance of these features in the generated summary of \emph{Heart of Darkness} is clear -- nearly every sentence contains one name, and the most important plot point captured is indeed one such life event  (``Mistah Kurtz -- he dead.'').

\begin{table}[ht]
\small
\begin{centering}
\begin{tabular}{|l|}
\hline
1. IS\_NAME\\ \hline 
2. DECILE\_0\\ \hline 
3. TF-IDF < 100\\ \hline 
4. DECILE\_8\\ \hline 
5. mr.\\ \hline 
6. TF-IDF < 10 \\ \hline
7. father\\ \hline
8. love \\ \hline 
9. son\\ \hline 
10. brother \\ \hline 
11. years\\ \hline 
12. young\\ \hline 
13. mother\\ \hline 
14. family \\ \hline 
15. DECILE\_9 \\ \hline 
16. daughter \\ \hline 
17. wife \\ \hline 
18. man \\ \hline
19. boy \\ \hline 
20. life \\ \hline 
21. death \\ \hline 
22. house \\ \hline 
23. chapter \\ \hline 
24. child \\ \hline 
25. sir \\ \hline 
\end{tabular}
\caption{\label{words} Strongest features predicting inclusion in a summary.}
\end{centering}
\end{table}

\section{Conclusion}

We present here two new methods optimized for aligning the full text of books with comparatively much shorter summaries, where the assumptions of the possibility of an exact word or phrase alignment may not always hold.  While these methods perform competitively in a downstream evaluation, book summarization clearly remains a challenging task.  Nevertheless, improved book/summary alignments hold intrinsic value in shedding light on what features of a work are deemed ``summarizable'' by human editors, and may potentially be exploited by tasks beyond summarization as well.

\appendix
{\small 
\section{Generated summary for \emph{Heart of Darkness}}

\begin{itemizesquish}
\item " And this also , " said Marlow suddenly , " has been one of the dark places of the earth . " He was the only man of us who still " followed the sea . " The worst that could be said of him was that he did not represent his class .
\item No one took the trouble to grunt even ; and presently he said , very slow -- " I was thinking of very old times , when the Romans first came here , nineteen hundred years ago -- the other day .... Light came out of this river since -- you say Knights ?
\item We looked on , waiting patiently -- there was nothing else to do till the end of the flood ; but it was only after a long silence , when he said , in a hesitating voice , " I suppose you fellows remember I did once turn fresh - water sailor for a bit , " that we knew we were fated , before the ebb began to run , to hear about one of Marlow ' s inconclusive experiences .
\item I know the wife of a very high personage in the Administration , and also a man who has lots of influence with , ' etc . She was determined to make no end of fuss to get me appointed skipper of a river steamboat , if such was my fancy .
\item He shook hands , I fancy , murmured vaguely , was satisfied with my French .
\item I found nothing else to do but to offer him one of my good Swede ' s 
\item Kurtz was ... I felt weary and irritable .
\item Kurtz was the best agent he had , an exceptional man , of the greatest importance to the Company ; therefore I could understand his anxiety .
\item I heard the name of Kurtz pronounced , then the words , ' take advantage of this unfortunate accident . ' One of the men was the manager .
\item Kurtz , ' I continued , severely , ' is General Manager , you won ' t have the opportunity . ' " He blew the candle out suddenly , and we went outside .
\item The approach to this Kurtz grubbing for ivory in the wretched bush was beset by as many dangers as though he had been an enchanted princess sleeping in a fabulous castle .
\item In a moment he came up again with a jump , possessed himself of both my hands , shook them continuously , while he gabbled : ' Brother sailor ... honour ... pleasure ... delight ... introduce myself ... Russian ... son of an arch - priest ... Government of Tambov ... What ?
\item Where ' s a sailor that does not smoke ? " " The pipe soothed him , and gradually I made out he had run away from school , had gone to sea in a Russian ship ; ran away again ; served some time in English ships ; was now reconciled with the arch - priest .
\item " He informed me , lowering his voice , that it was Kurtz who had ordered the attack to be made on the steamer .
\item " We had carried Kurtz into the pilot - house : there was more air there .
\item Suddenly the manager ' s boy put his insolent black head in the doorway , and said in a tone of scathing contempt : " ' Mistah Kurtz -- he dead . ' " All the pilgrims rushed out to see .
\item That is why I have remained loyal to Kurtz to the last , and even beyond , when a long time after I heard once more , not his own voice , but the echo of his magnificent eloquence thrown to me from a soul as translucently pure as a cliff of crystal .
\item Kurtz ' s knowledge of unexplored regions must have been necessarily extensive and peculiar -- owing to his great abilities and to the deplorable circumstances in which he had been placed : therefore -- ' I assured him Mr .
\item ' There are only private letters . ' He withdrew upon some threat of legal proceedings , and I saw him no more ; but another fellow , calling himself Kurtz ' s cousin , appeared two days later , and was anxious to hear all the details about his dear relative ' s last moments .
\item Incidentally he gave me to understand that Kurtz had been essentially a great musician .
\item I had no reason to doubt his statement ; and to this day I am unable to say what was Kurtz ' s profession , whether he ever had any -- which was the greatest of his talents .
\item This visitor informed me Kurtz ' s proper sphere ought to have been politics ' on the popular side . ' He had furry straight eyebrows , bristly hair cropped short , an eyeglass on a broad ribbon , and , becoming expansive , confessed his opinion that Kurtz really couldn ' t write a bit -- ' but heavens ! how that man could talk .
\item All that had been Kurtz ' s had passed out of my hands : his soul , his body , his station , his plans , his ivory , his career .
\item And , by Jove ! the impression was so powerful that for me , too , he seemed to have died only yesterday -- nay , this very minute .
\item He had given me some reason to infer that it was his impatience of comparative poverty that drove him out there . " ' ... Who was not his friend who had heard him speak once ? ' she was saying .
\item Would they have fallen , I wonder , if I had rendered Kurtz that justice which was his due ?
\end{itemizesquish}
}

\clearpage
\normalsize
\bibliographystyle{acl}
\bibliography{newbibliography,more}

\end{document}